\documentclass{article}
\PassOptionsToPackage{authoryear}{natbib}   
\usepackage[preprint]{neurips_2025}

\title{Think, Remember, Navigate: Zero-Shot Object-Goal Navigation with VLM-Powered Reasoning}
\workshoptitle{NeurIPS 2025 Workshop on SPACE in Vision, Language, and Embodied AI (SpaVLE)}

\usepackage[utf8]{inputenc} % allow utf-8 input
\usepackage[T1]{fontenc}    % use 8-bit T1 fonts
\usepackage{hyperref}       % hyperlinks
\usepackage{url}            % simple URL typesetting
\usepackage{booktabs}       % professional-quality tables
\usepackage{amsfonts}       % blackboard math symbols
\usepackage{nicefrac}       % compact symbols for 1/2, etc.
\usepackage{microtype}      % microtypography
\usepackage{xcolor}         % colors
\usepackage{graphicx}       % figures
\usepackage{amsmath}
\usepackage{subfig}         % for subfigures
\usepackage{algorithm}
\usepackage{algorithmic}
\usepackage{amssymb}
\usepackage{tabularx}
\usepackage{placeins}
\usepackage{listings}

\hypersetup{
    colorlinks=true,
    linkcolor=blue,
    citecolor=blue,
    urlcolor=blue,
    breaklinks=true
}

\author{
  Mobin Habibpour \qquad Fatemeh Afghah\\
  Holcombe Department of Electrical and Computer Engineering\\
  Clemson University, Clemson, SC, USA\\
  \texttt{mhabibp@clemson.edu} \quad \texttt{fafghah@clemson.edu}
}

\begin{document}
%%%%%%%%%%%%%%%%%%%%%%%%%%%%%%%%%%%%%%%%%%%%%%%%%%%%%%%%%%%%%%%%

\maketitle

% --- Abstract ---
\begin{abstract}
While Vision-Language Models (VLMs) are set to transform robotic navigation, existing methods often underutilize their reasoning capabilities. To unlock the full potential of VLMs in robotics, we shift their role from passive observers to active strategists in the navigation process. Our framework outsources high-level planning to a VLM, which leverages its contextual understanding to guide a frontier-based exploration agent. This intelligent guidance is achieved through a trio of techniques: structured chain-of-thought prompting that elicits logical, step-by-step reasoning; dynamic inclusion of the agent's recent action history to prevent getting stuck in loops; and a novel capability that enables the VLM to interpret top-down obstacle maps alongside first-person views, thereby enhancing spatial awareness. When tested on challenging benchmarks like HM3D, Gibson, and MP3D, this method produces exceptionally direct and logical trajectories, marking a substantial improvement in navigation efficiency over existing approaches and charting a path toward more capable embodied agents.
\end{abstract}

% --- Main Content ---
\section{Introduction}
\label{sec:intro}

A primary challenge in robotics is enabling autonomous navigation in unknown environments, a skill crucial for tasks like search and rescue or industrial inspection. Object Goal Navigation (ObjectNav), where an agent must locate a specific object in a new setting, is a particularly demanding version of this problem \cite{chaplot2020object, sun2024survey}. It requires a combination of advanced spatial awareness and semantic comprehension. Traditional approaches, which often depend on geometric maps and predetermined planning strategies, have difficulty generalizing to novel environments and thus fall short of achieving genuinely intelligent exploration.

The emergence of powerful Vision-Language Models (VLMs) has opened a new frontier, giving robots the potential to interpret their surroundings with a human-like understanding of context. Despite this promise, VLMs have typically been integrated into navigation systems in a limited capacity. Many current methods assign the VLM a passive function, such as scene description or query answering, rather than making it the primary strategist. This underuse is a result of superficial knowledge integration, inflexible prompting techniques, and a lack of memory, all of which lead to inefficient navigation and a restricted role for the VLM.

To overcome these shortcomings, we propose a novel framework that delegates high-level planning to a VLM, using its inherent contextual knowledge to direct navigation. Our approach reimagines the VLM's function, elevating it from a simple reactive component to the main navigator. By combining the VLM's emergent planning skills with frontier-based exploration, dynamic prompting, and multi-view fusion, our framework achieves robust and efficient navigation without relying on conventional planners. Our main contributions include:

\begin{itemize}
\item \textbf{Chain-of-Thought (CoT) for Navigation:} We employ CoT reasoning within the VLM, which allows it to produce more logical and context-aware instructions by methodically thinking through each step of the navigation task.

\item \textbf{Dynamic Prompts with Action History:} Our system uses advanced prompts that include the agent’s recent actions. This helps to avoid common issues like getting stuck in loops or indecisive movements, leading to more reliable exploration.

\item \textbf{Top-Down Map Interpretation:} We enhance the VLM’s reasoning by enabling it to analyze top-down obstacle maps in conjunction with its first-person view, giving it a better sense of the overall space for long-term planning.
\end{itemize}

\section{Related Work}
\label{sec:related}

The challenge of Object Goal Navigation (ObjectNav) has been approached from several angles, beginning with end-to-end learning frameworks that often struggled to generalize beyond their training data \cite{mousavian2019visual, ye2021efficient}. Modular pipelines emerged as an alternative, deconstructing the problem into perception, mapping, and planning stages \cite{chaplot2020object}. While more robust, these systems could be brittle and prone to cascading errors between components \cite{kumar2021gcexp}. This has motivated a recent surge in zero-shot methodologies that harness the world knowledge of large pre-trained models, thereby avoiding the need for extensive task-specific fine-tuning \cite{sun2024survey}.

The integration of Vision-Language Models (VLMs) first involved using them as powerful feature extractors, with methods like CLIP on Wheels (CoWs) \cite{gadre2022clip} leveraging embeddings to link visual scenes with a target object's name. The role of language models soon became more active, with systems like ESC applying common-sense reasoning to guide exploration \cite{zhou2023esc}. More recently, the frontier of ObjectNav has shifted toward offloading high-level strategy entirely to large models. This paradigm includes diverse techniques such as translating maps into text for an LLM to score exploration paths (L3MVN \cite{yu2023l3mvn}), employing a VLM to directly evaluate the semantic promise of frontiers (VLFM \cite{yokoyama2024vlfm}), or even using a VLM to imagine and select optimal future viewpoints (ImagineNav \cite{zhao2024imaginenav}).

Our work advances this paradigm by enhancing the VLM's cognitive role in navigation. We introduce a framework that empowers the VLM with a more profound understanding of the task through a synergistic combination of three key techniques: structured Chain-of-Thought (CoT) prompting to elicit more logical, step-by-step analysis \cite{wei2023chainofthoughtprompting}; the incorporation of a memory of recent actions to prevent stagnation; and a novel method for providing multimodal spatial context by enabling the VLM to interpret top-down obstacle maps in conjunction with its egocentric view. This holistic approach results in a more effective zero-shot navigator capable of generating more coherent and efficient trajectories.

\section{Methodology}
\label{sec:method}

\begin{figure*}[t]
    \centering
    \includegraphics[width=\textwidth]{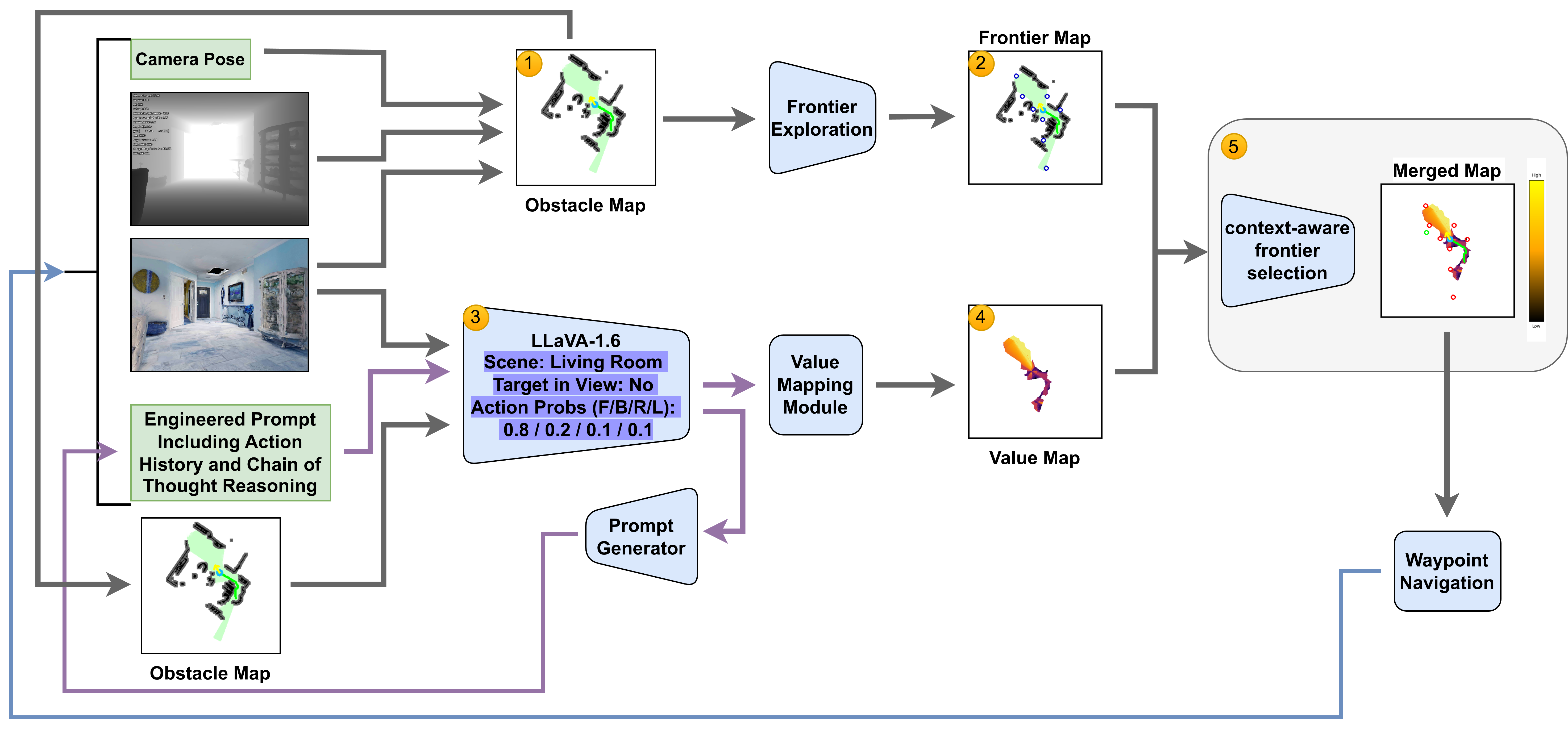}
    \caption{Our System Pipeline. (1) Sensor data is used to create an Obstacle Map. (2) Geometric frontiers are detected on this map. (3) The LLaVA-1.6 VLM analyzes the agent's egocentric view, the map, and a dynamic prompt that includes action history. (4) The VLM produces semantic scores, which are then used to build a Value Map that indicates the relevance of different areas. (5) The Frontier and Value Maps are combined to prioritize waypoints, directing the agent toward the most promising regions.}
    \label{fig:pipeline}
\end{figure*}

Our method's navigation is organized into three phases: initialization, exploration, and goal navigation. Our primary contribution is in the exploration phase, where we use a VLM to bring contextual intelligence to a standard frontier-based exploration framework. As shown in Figure~\ref{fig:pipeline}, the VLM processes the agent’s first-person view, a top-down obstacle map, and a specially designed prompt. It then serves as a high-level guide, suggesting the next move based on its comprehensive understanding of the scene and the goal. Algorithm~\ref{alg:object_nav_explore} outlines the comprehensive procedure for the exploration stage, detailing how the VLM guidance, Value Map updates, and frontier prioritization are integrated to navigate the environment.

\subsection{Waypoint Generation via Frontier Exploration}
\label{way} 
The agent is equipped with a 2D action space, including moving forward, turning, and stopping. We use the VER algorithm \cite{wijmans2022ver} for low-level motion control to guide the agent toward specific 2D waypoints. These waypoints are identified through a frontier-based exploration method \cite{topiwala2018frontierbasedexplorationautonomous}, which analyzes depth data to create an obstacle map and find the boundaries between explored and unexplored areas. The midpoints of these frontiers are chosen as potential targets for exploration, promoting a thorough search of the environment.

\subsection{Value Map Creation using VLM Guidance}
\label{value}
\begin{figure}[!t]
    \centering
    \subfloat[]{\includegraphics[width=0.4\columnwidth, height=4cm, keepaspectratio]{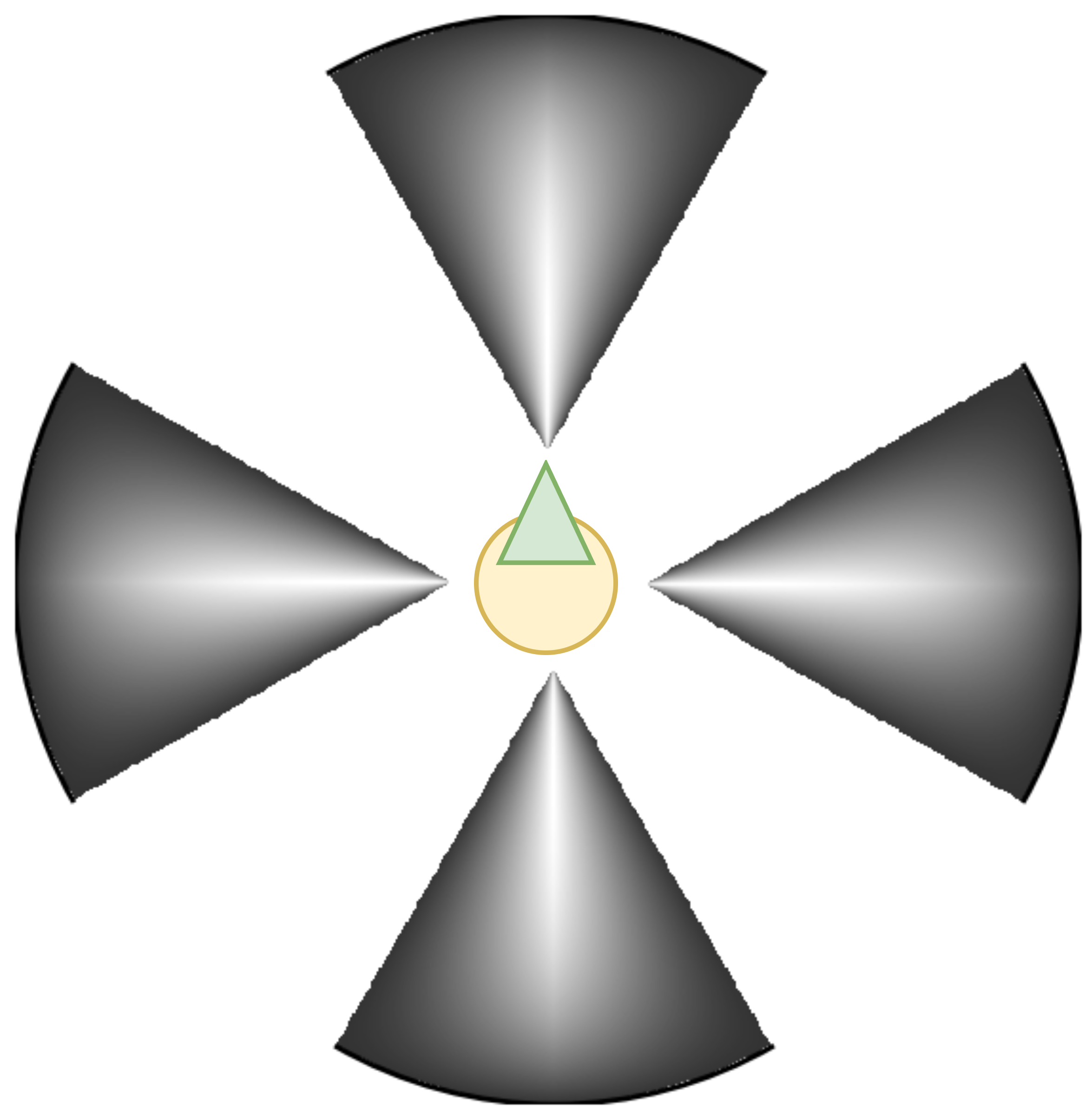}% Assuming file exists
    \label{fig:fov}}
    \hfil
    \subfloat[]{\includegraphics[width=0.4\columnwidth, height=4cm, keepaspectratio]{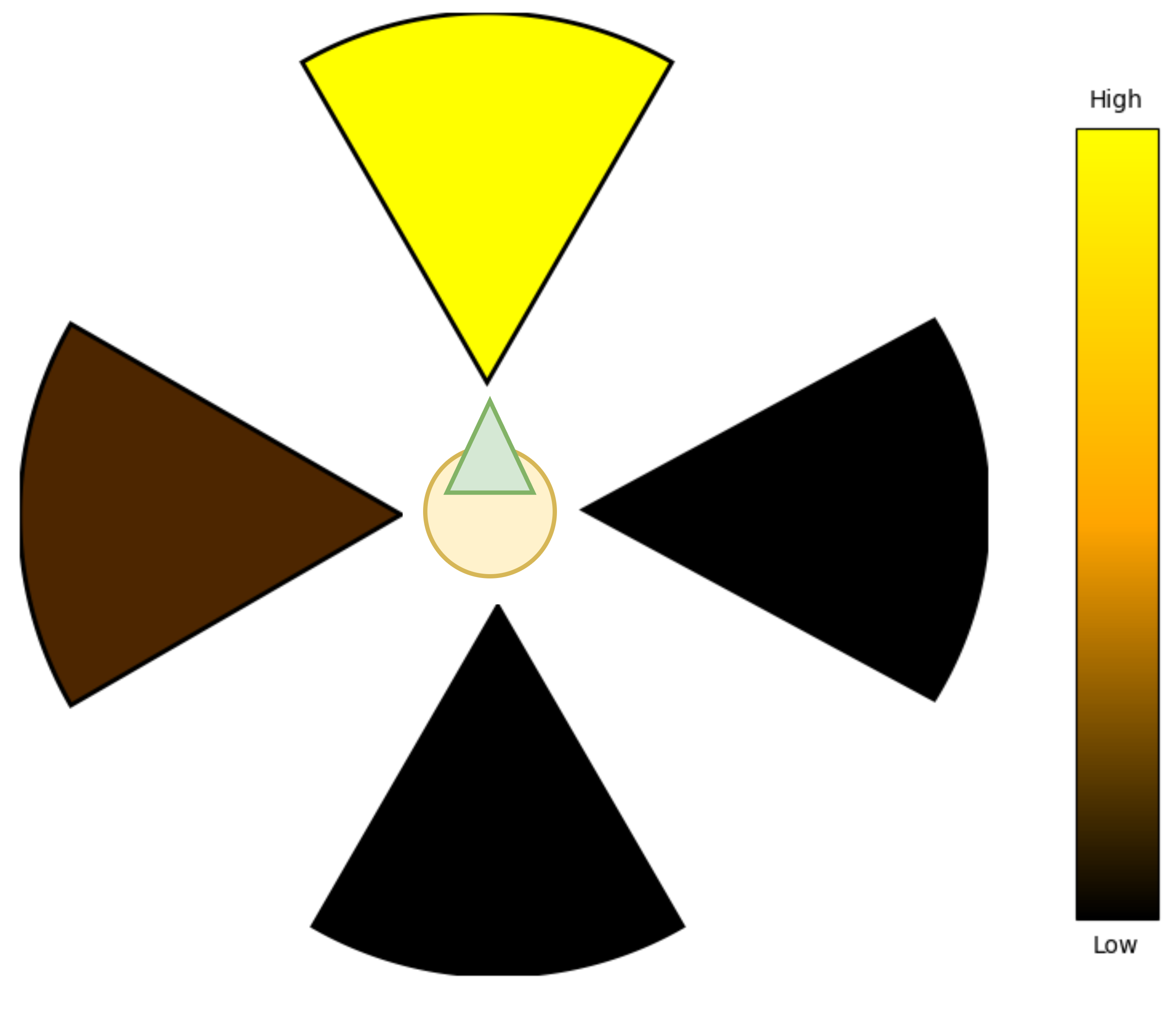}% Assuming file exists
    \label{fig:actions}}
    \caption{Conceptual illustration of the value map components. (a) Robot's field of view (FOV) and the associated viewing uncertainty cone. (b) Example action space visualization with VLM-assigned scores: [Forward: 0.9, Backward: 0, Right: 0, Left: 0.1].}
    \label{fig:scoreshape}
\end{figure}

Since geometric frontiers do not have semantic meaning, we use the VLM to assign probability scores to actions like moving forward or turning left, based on the current view, action history, and our prompt. These scores are used to build a \textit{value map} that reflects how promising different areas are. The scores are projected onto the local map and are adjusted by a \textit{viewing uncertainty} factor, which accounts for the fact that the VLM's judgments may be less accurate for distant or peripheral regions, as conceptually depicted in Figure~\ref{fig:scoreshape}. The confidence level $c$ for a point at a distance $d$ and angle $\theta$ is calculated as:
\begin{equation}
c(d, \theta) = e^{-\lambda d} \cdot \cos^2\left( \frac{\theta}{\theta_{\text{fov}}/2} \cdot \frac{\pi}{2} \right)
\label{eq:confidence}
\end{equation}

Here, $\theta_{\text{fov}}$ is the camera's field of view, and $\lambda$ is a parameter for distance decay. When the agent observes areas that overlap with previously seen regions, the semantic value $v^{\text{new}}_{i,j}$ and confidence $c^{\text{new}}_{i,j}$ for each pixel $(i,j)$ are updated through a confidence-weighted average. This ensures that more reliable observations have a stronger influence:
\begin{equation}
v^{\text{new}}_{i,j} = \frac{c^{\text{curr}}_{i,j} v^{\text{curr}}_{i,j} + c^{\text{prev}}_{i,j} v^{\text{prev}}_{i,j}}{c^{\text{curr}}_{i,j} + c^{\text{prev}}_{i,j}}
\quad \text{and} \quad
c^{\text{new}}_{i,j} = \frac{(c^{\text{curr}}_{i,j})^2 + (c^{\text{prev}}_{i,j})^2}{c^{\text{curr}}_{i,j} + c^{\text{prev}}_{i,j}}
\label{eq:value_update}
\end{equation}
The resulting value map is then combined with the geometric frontier map, enabling the system to prioritize exploration of the most semantically relevant areas.
\subsection{Top-Down View Map Parsing by VLM} \label{map}
\begin{figure}[h]
    \centering
    \subfloat[Top-Down Obstacle Map]{\includegraphics[width=0.4\columnwidth]{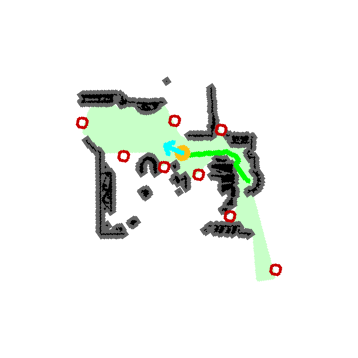}\label{fig:Obstacle}}
    \hfill
    \subfloat[Egocentric View]{\includegraphics[width=0.4\columnwidth]{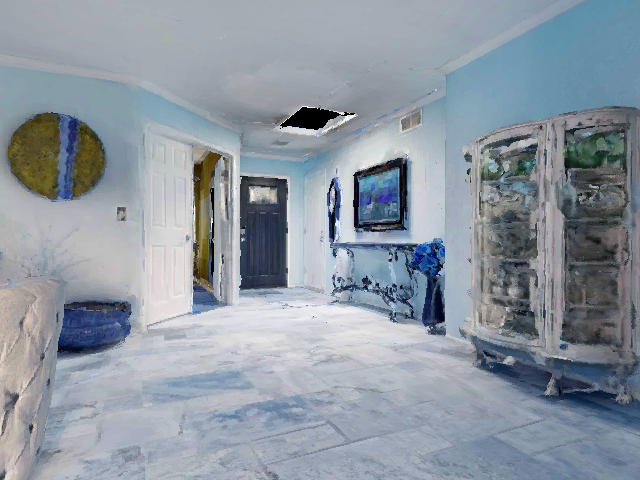}\label{fig:EgoView}}
    \caption{Dual visual inputs for the VLM: (a) the top-down map shows the spatial layout with obstacles (in gray) and the agent's heading (arrow), while (b) the egocentric view provides a first-person perspective. This combination improves the VLM's spatial understanding.}
    \label{fig:navigation}
\end{figure}

Drawing inspiration from human spatial reasoning, our framework equips the VLM with both a first-person egocentric view and a top-down obstacle map (Figure \ref{fig:navigation}). Although VLMs are not typically trained on map data, we provide clear instructions in the prompt, such as: \textit{"The second image is a top-down obstacle map..."}. This merging of first-person and overhead perspectives enhances the VLM's spatial awareness, leading to better-informed navigation choices, a finding supported by our ablation studies.

Our choice of VLM was guided by a balance of reasoning capability and computational efficiency. After evaluating several open-source models, we selected LLaVA-1.6 (7B) \cite{liu2024improvedbaselinesvisualinstruction} as it provided strong performance on multi-step reasoning tasks while maintaining an inference speed suitable for our navigation loop. During the final goal navigation phase, our system employs a robust hybrid strategy to mitigate potential failures in object detection. If an object is detected with high confidence by standard detectors (e.g., YOLOv7, Grounding-DINO), the system verifies the finding with the VLM and uses a lightweight segmentation model (Mobile-SAM) to delineate the object's boundaries, ensuring a more reliable final approach to the target.

\section{Prompt Engineering}
\label{sec:prompt}
Successful interaction with the VLM depends heavily on the design of the prompt. We use several techniques to obtain structured and logical responses that are useful for navigation. CoT prompting improves a model's reasoning by guiding it through a series of intermediate steps to solve a complex problem \cite{wei2023chainofthoughtprompting}. Instead of just asking for the best move, our prompt (see listing \ref{lst:prompt_full_cot}) leads the VLM through a logical sequence:
(1) \textit{Determine the room type.} (2) \textit{Evaluate if the target object is likely to be in that room.} (3) \textit{Confirm if the target is currently visible.} (4) \textit{Suggest the most sensible action and assign scores.}
This methodical approach, which was refined through extensive testing, ensures that the VLM's understanding of the scene is consistent with its recommended actions, reducing the chances of logical errors.

\begin{figure}[!t]
    \centering
    \captionsetup[subfigure]{font=footnotesize} % Ensure subfigure captions are small
     \subfloat[]{\includegraphics[width=0.45\columnwidth, height=4cm, keepaspectratio]{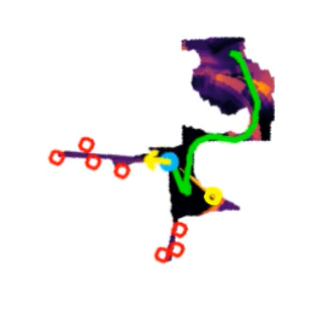}% Assuming file exists
     \label{fig:view1_loop}} % Changed label to avoid conflict
     \hfil
     \subfloat[]{\includegraphics[width=0.45\columnwidth, height=4cm, keepaspectratio]{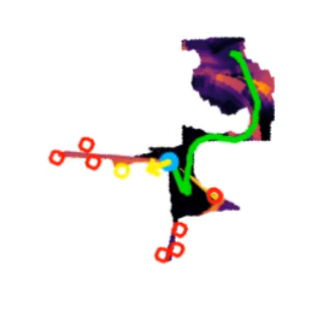}% Assuming file exists
     \label{fig:view2_loop}} % Changed label to avoid conflict
    \caption{\footnotesize Example of potential decision loop. Without action history, oscillating evaluations between points (a) and (b) could cause stagnation. Tracking history helps break such cycles.}
    \label{fig:decision_loop}
\end{figure}

A common issue in navigation is when an agent gets stuck in a loop or becomes indecisive, especially in visually similar environments like long hallways, a scenario illustrated in Figure \ref{fig:decision_loop}. To address this, we keep a record of the agent’s last 10 actions. This history is included in the VLM prompt with instructions such as: \textit{"Maintain forward progress... and avoid repetitive actions."} If the agent still gets stuck, a backup plan is activated, which temporarily repeats the last successful action to break the cycle. Our ablation study in Section \ref{sec:ablation} confirms that this memory-based strategy is essential for reliable long-term navigation.

\section{Experimental Evaluation}
\label{sec:result}
We tested our framework in the Habitat simulator \cite{habitatchallenge2023}, using the HM3D, Gibson, and MP3D ObjectNav benchmarks \cite{ramakrishnan2021habitat, xia2018gibsonenvrealworldperception, Matterport3D}. The main metrics for evaluation were Success weighted by Path Length (SPL) and Success Rate (SR) \cite{anderson2018evaluation}. We compared our approach with other top zero-shot methods. As Table \ref{tab:main_results} shows, our approach delivers exceptional efficiency, achieving the highest SPL on the demanding HM3D and MP3D datasets, and strong performance on Gibson. This indicates that our method's deep integration of VLM reasoning leads to more direct and intentional paths. This high efficiency is a direct result of our CoT and action history techniques, which prevent aimless exploration and keep the agent on a more logical course.

\begin{table}[h]
\caption{Performance Comparison on ObjectNav Datasets. Our approach utilizes the LLaVA-1.6 (7B) model. SR and SPL are reported in percent. Top values are in bold.}
\label{tab:main_results}
\centering
\small
\begin{tabular}{@{}lcccccc@{}}
\toprule
\textbf{Method} & \multicolumn{2}{c}{\textbf{HM3D}} & \multicolumn{2}{c}{\textbf{MP3D}} & \multicolumn{2}{c}{\textbf{Gibson}} \\
& \textbf{SR} & \textbf{SPL} & \textbf{SR} & \textbf{SPL} & \textbf{SR} & \textbf{SPL} \\
\midrule
CoW \cite{Gadre2022CLIPOW} & - & 6.3 & 3.7 & 7.4 & - & - \\
ZSON \cite{majumdar2023zsonzeroshotobjectgoalnavigation} & 25.5 & 12.6 & 15.3 & 4.8 & - & - \\
ESC \cite{zhou2023esc} & 39.2 & 22.3 & 28.7 & 14.2 & - & - \\
VLFM \cite{yokoyama2024vlfm} & 52.5 & 30.4 & \textbf{36.4} & 17.5 & 84.0 & 52.2  \\
L3MVN \cite{yu2023l3mvn} & 50.4 & 23.1 & - & - & 76.1 & 37.7 \\
GAMap \cite{yuan2024gamapzeroshotobjectgoal} & 53.1 & 26.0 & - & - & \textbf{85.7} & \textbf{55.5} \\
\midrule
\textbf{Our Approach} & \textbf{54.3} & \textbf{31.1} & 36.0 & \textbf{17.7} & 80.2 & 53.0 \\
\bottomrule
\end{tabular}%
\end{table}

\subsection{Ablation Studies}
\label{sec:ablation}
To determine the impact of each part of our system, we conducted a series of ablation studies on a smaller set of benchmark episodes (Table \ref{tab:ablation}). The findings show that every component adds value. Removing the top-down map resulted in a small drop in performance, while taking out the Chain-of-Thought reasoning caused a more significant decrease. Notably, the largest decline in performance happened when the \textbf{action history module} was removed, with the Success Rate on HM3D plummeting from 54.3\% to 44.0\%. This severe drop underscores that providing the agent with a short-term memory of its recent path is a necessity for robust, long-horizon navigation. Without this temporal context, the agent is prone to re-evaluating the same frontiers and becoming trapped in unproductive, oscillatory loops—a common failure mode observed in visually ambiguous environments.

\begin{table}[h]
\caption{Results of the Ablation Study on a 50-episode subset. Each row indicates performance when the corresponding component is removed. The removal of action history has the most negative effect.}
\label{tab:ablation}
\centering
\setlength{\tabcolsep}{4pt}
\begin{tabular}{@{}lcccccc@{}}
\toprule
\textbf{Ablation Condition} & \multicolumn{2}{c}{\textbf{HM3D}} & \multicolumn{2}{c}{\textbf{MP3D}} & \multicolumn{2}{c}{\textbf{Gibson}} \\
& \textbf{SR (\%)} & \textbf{SPL (\%)} & \textbf{SR (\%)} & \textbf{SPL (\%)} & \textbf{SR (\%)} & \textbf{SPL (\%)} \\
\midrule
\textbf{Full Framework} & \textbf{54.3} & \textbf{31.1} & \textbf{36.0} & \textbf{17.7} & \textbf{80.2} & \textbf{53.0} \\
\addlinespace 
Without Chain-of-Thought & 51.2 & 29.0 & 33.9 & 16.5 & 75.6 & 49.4 \\
\addlinespace
Without Action History & 44.0 & 23.7 & 29.2 & 13.5 & 65.0 & 40.4 \\
\addlinespace
Without Obstacle Map & 53.6 & 29.6 & 35.5 & 16.9 & 79.2 & 50.5 \\
\bottomrule
\end{tabular}
\end{table}

\subsubsection{Effectiveness of Chain-of-Thought Prompting}
To validate our structured reasoning approach, we tested prompts with varying levels of CoT complexity, based on the hypothesis that guiding the VLM through a logical sequence of questions would yield more coherent and effective navigation decisions. We evaluated four configurations ranging in complexity: a baseline \textbf{No CoT} prompt that only requested action scores without any reasoning; a \textbf{Basic CoT} prompt asking for the single best action to find the target; an \textbf{Intermediate CoT} prompt that added a query for scene identification (e.g., "what room are you in?") before asking for an action; and our \textbf{Full NaviGen CoT}, a complete, multi-step prompt that requires the VLM to first identify the scene, then assess the likelihood of finding the target object there, and finally recommend an action based on this analysis.

The results, shown in Table \ref{tab:cot_ablation}, confirm a direct positive correlation between the depth of the CoT prompt and navigation performance. For instance, on HM3D, progressing from a non-CoT prompt to our full framework improved the SR from 51.2\% to 54.3\% and boosted the efficiency (SPL) from 29.0 to 31.1. This shows that compelling the VLM to "think step-by-step" is a functional requirement for translating visual input into successful, goal-oriented action.

\begin{table}[h]
\caption{Ablation Study on Chain-of-Thought Prompt Complexity. Performance evaluated on a 50-episode subset.}
\label{tab:cot_ablation}
\centering
\small
\begin{tabular}{@{}lcccccc@{}}
\toprule
\textbf{Prompt Variation} & \multicolumn{2}{c}{\textbf{HM3D}} & \multicolumn{2}{c}{\textbf{MP3D}} & \multicolumn{2}{c}{\textbf{Gibson}} \\
& \textbf{SR} & \textbf{SPL} & \textbf{SR} & \textbf{SPL} & \textbf{SR} & \textbf{SPL} \\
\midrule
No CoT (scores only, \ref{lst:prompt_no_cot}) & 51.2 & 29.0 & 32.5 & 15.1 & 75.4 & 49.5 \\
\addlinespace 
Basic CoT ("Best action?", \ref{lst:prompt_basic_cot}) & 52.0 & 29.5 & 33.1 & 15.9 & 77.0 & 50.3 \\
\addlinespace
Intermediate CoT (Scene ID, \ref{lst:prompt_intermediate_cot}) & 53.3 & 30.2 & 34.8 & 16.8 & 78.9 & 51.7 \\
\addlinespace
\textbf{Full CoT (multi-step, \ref{lst:prompt_full_cot})} & \textbf{54.3} & \textbf{31.1} & \textbf{36.0} & \textbf{17.7} & \textbf{80.2} & \textbf{53.0} \\
\bottomrule
\end{tabular}
\end{table}

\begin{figure*}[!t]
    \centering
    % Set column separation to a small value for a tight layout.
    \setlength{\tabcolsep}{2pt}
    % A table will structure the entire figure.
    \begin{tabular}{c c}
        % --- TOP HEADER ROW ---
        % Add column titles. The first cell is empty to align with the row labels.
        & 
        \begin{tabular}{cccc}
            \multicolumn{1}{p{0.25\textwidth}}{\centering\small\textbf{Step 100}} &
            \multicolumn{1}{p{0.25\textwidth}}{\centering\small\textbf{Step 200}} &
            \multicolumn{1}{p{0.25\textwidth}}{\centering\small\textbf{Step 300}} &
            \multicolumn{1}{p{0.25\textwidth}}{\centering\small\textbf{Final Trajectory}}
        \end{tabular} \\ % END OF HEADER ROW, REDUCED SPACE

        % --- ROW 1: NO CoT, EPISODE 1 ---
        \rotatebox{90}{\parbox{2.5cm}{\centering\small\textbf{No CoT}}} &
        % A nested table holds all content for this row.
        \begin{tabular}{cccc}
            \begin{minipage}[t]{0.25\textwidth}\includegraphics[width=\linewidth, height=2.5cm, keepaspectratio]{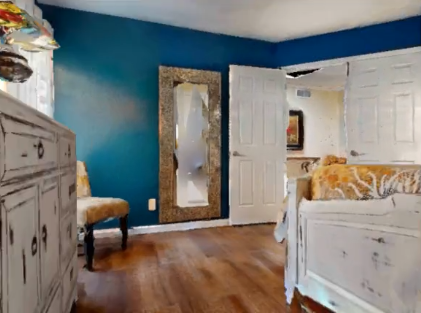}\begin{lstlisting}[basicstyle=\ttfamily\tiny\bfseries, breakindent=0pt]
It appears that the TV is located to the right of the cabinet. Therefore, the best action for the robot to take is to turn right.
Scores: F:0.0, B:0.0, L:0.0, R:1.0
            \end{lstlisting}\end{minipage} &
            \begin{minipage}[t]{0.25\textwidth}\includegraphics[width=\linewidth, height=2.5cm, keepaspectratio]{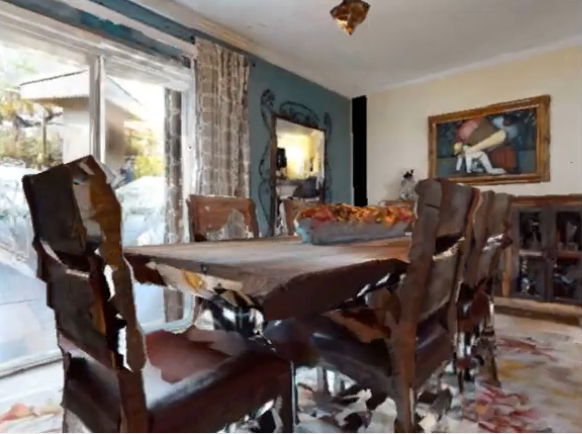}\begin{lstlisting}[basicstyle=\ttfamily\tiny\bfseries, breakindent=0pt]
It appears that the TV is located in the upper right corner of the room. To navigate to the TV, the best action would be to turn right.
Scores: F:0.0, B:0.0, L:0.0, R:1.0
            \end{lstlisting}\end{minipage} &
            \begin{minipage}[t]{0.25\textwidth}\includegraphics[width=\linewidth, height=2.5cm, keepaspectratio]{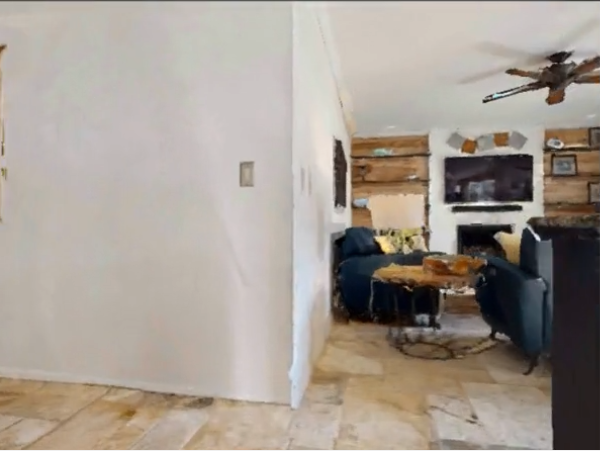}\begin{lstlisting}[basicstyle=\ttfamily\tiny\bfseries, breakindent=0pt]
It appears that the robot is already in the room where the TV is located. The TV is visible in the background, to the left of the image. To reach the TV, the robot should go forward.
Scores: F:1.0, B:0.0, L:0.0, R:0.0
            \end{lstlisting}\end{minipage} &
            \begin{minipage}[t]{0.25\textwidth}\includegraphics[width=\linewidth, height=2.5cm, keepaspectratio]{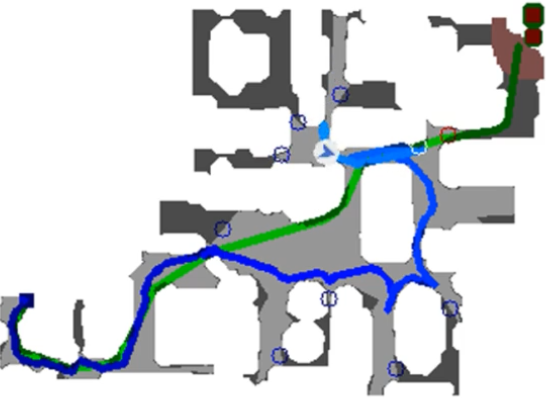}\end{minipage}
        \end{tabular} \\[-4pt] % END OF ROW 1, REDUCED SPACE

        % --- ROW 2: FULL CoT, EPISODE 1 ---
        \rotatebox{90}{\parbox{2.2cm}{\centering\small\textbf{Full CoT}}} &
        \begin{tabular}{cccc}
        
            \begin{minipage}[t]{0.25\textwidth}\includegraphics[width=\linewidth, height=2.5cm, keepaspectratio]{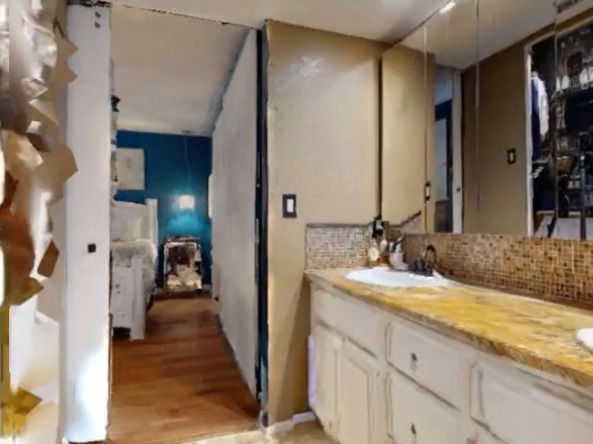}\begin{lstlisting}[basicstyle=\ttfamily\tiny\bfseries, breakindent=0pt]
Part of the House: Bathroom
Can a TV be Found Here?: No
Recommended Action: Go forward
Reasoning: The most logical next action for the robot is to continue moving forward to exit the bathroom.
Scores: F:0.8, B:0.2, L:0.0, R:0.0
            \end{lstlisting}\end{minipage} &
            \begin{minipage}[t]{0.25\textwidth}\includegraphics[width=\linewidth, height=2.5cm, keepaspectratio]{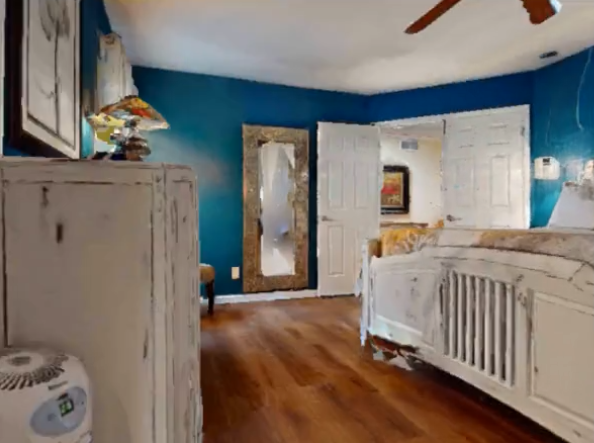}\begin{lstlisting}[basicstyle=\ttfamily\tiny\bfseries, breakindent=0pt]
Part of the House: Bedroom 
Can a TV be Found Here?: Yes
Recommended Action: Go forward
Reasoning: The most logical next action is to continue exploring the bedroom area.
Scores: F:0.9, B:0.1, L:0.0, R:0.0
            \end{lstlisting}\end{minipage} &
            \begin{minipage}[t]{0.25\textwidth}\includegraphics[width=\linewidth, height=2.5cm, keepaspectratio]{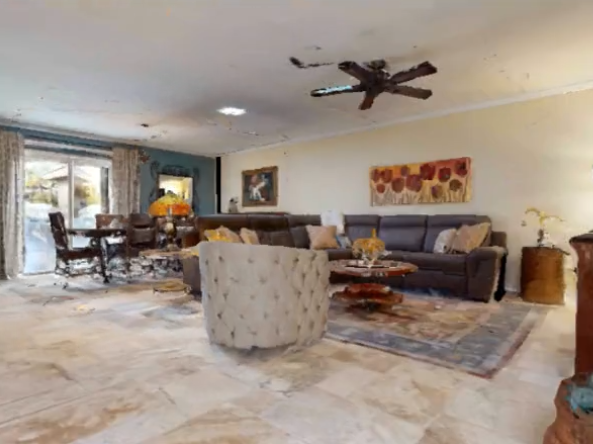}\begin{lstlisting}[basicstyle=\ttfamily\tiny\bfseries, breakindent=0pt]
Part of the House: Living room
Can a TV Be Found Here?: Yes
Recommended Action: Go forward
Reasoning: Since the TV has been found, the most logical next action is to move forward.
Scores: F:1.0, B:0.0, L:0.0, R:0.0
            \end{lstlisting}\end{minipage} &
            \begin{minipage}[t]{0.25\textwidth}\includegraphics[width=\linewidth, height=2.5cm, keepaspectratio]{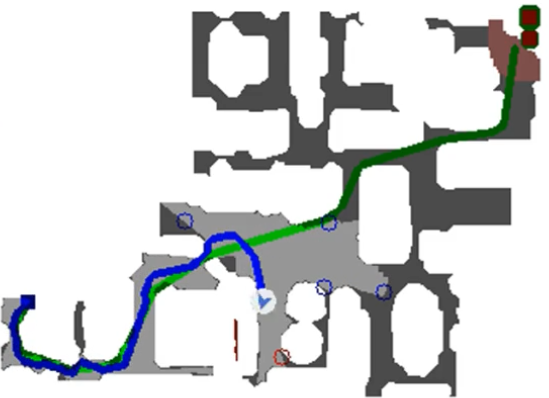}\end{minipage}
        \end{tabular} \\[-4pt] % END OF ROW 2, REDUCED SPACE
    
    \end{tabular}
    \caption{Qualitative analysis of navigation with and without our full CoT framework. The agent's view and the VLM's reasoning are shown at various timesteps. The \textbf{top row (No CoT)} displays an agent with basic reasoning that wanders aimlessly and fails to locate the target. The \textbf{bottom row (Full CoT)} illustrates how a structured, step-by-step reasoning process (such as identifying a bathroom, realizing a TV is not there, and choosing to leave) results in a more intelligent exploration strategy and a direct, successful path. This comparison underscores the crucial role of CoT in achieving more effective and intelligent navigation.}
    \label{fig:cot_qualitative}
\end{figure*}

\subsection{Qualitative Analysis and Limitations}

To investigate the reasons for our framework's efficiency, we conducted a qualitative study of the VLM's decision-making. Figure \ref{fig:cot_qualitative} offers a direct comparison between an agent using our full, multi-step Chain-of-Thought (CoT) prompt and a basic agent with a simpler, non-CoT prompt. This comparison clearly demonstrates how structured thinking leads to better navigation.

\begin{figure*}[!tp]
    \centering
    % Set column separation to zero to remove space between images
    \setlength{\tabcolsep}{0pt}
    % Define the table structure.
    \begin{tabular}{c c c c c c}
        % --- COLUMN HEADERS ---
        & \multicolumn{1}{p{0.19\textwidth}}{\centering\small\textbf{Step 10}} 
        & \multicolumn{1}{p{0.19\textwidth}}{\centering\small\textbf{Step 100}} 
        & \multicolumn{1}{p{0.19\textwidth}}{\centering\small\textbf{Step 150}} 
        & \multicolumn{1}{p{0.19\textwidth}}{\centering\small\textbf{Step 200}} 
        & \multicolumn{1}{p{0.19\textwidth}}{\centering\small\textbf{Step 300}} \\[-15pt] % Reduced space after header
        
        % --- ROW 1: Episode 1 (No Action History) ---
        \rotatebox{90}{\parbox{2.5cm}{\centering\small Episode 1 \\ (No History)}} &
        % Image cells
        \includegraphics[width=0.19\textwidth, height=2.5cm, keepaspectratio]{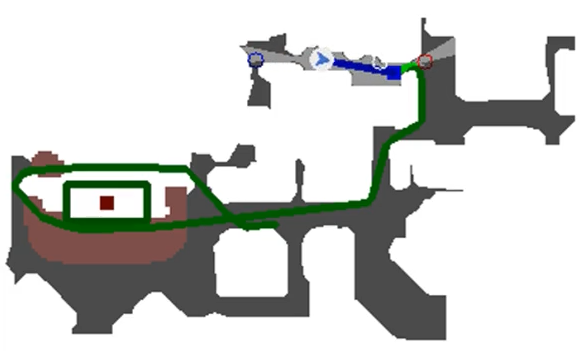} &
        \includegraphics[width=0.19\textwidth, height=2.5cm, keepaspectratio]{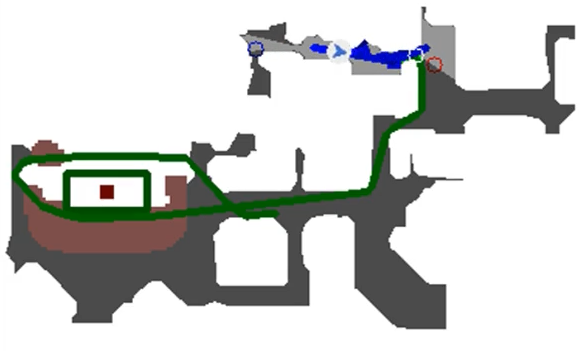} &
        \includegraphics[width=0.19\textwidth, height=2.5cm, keepaspectratio]{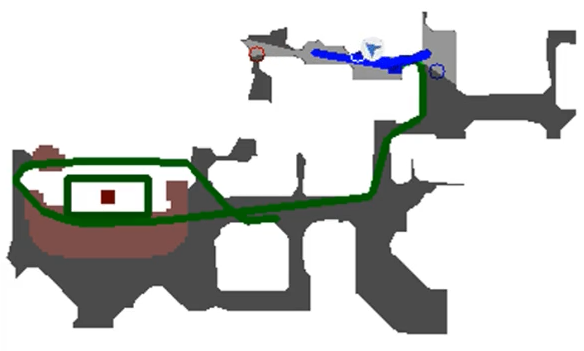} &
        \includegraphics[width=0.19\textwidth, height=2.5cm, keepaspectratio]{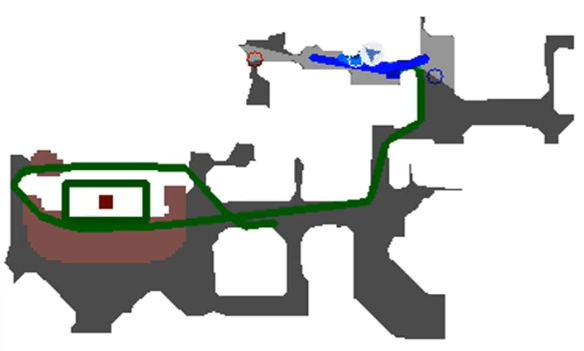} &
        \includegraphics[width=0.19\textwidth, height=2.5cm, keepaspectratio]{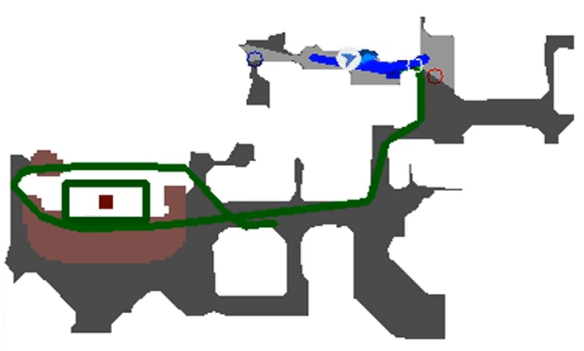} \\[-20pt] % Reduced space after row 1

        % --- ROW 2: Episode 1 (With Action History) ---
        \rotatebox{90}{\parbox{2.5cm}{\centering\small Episode 1 \\ (With History)}} &
        \includegraphics[width=0.19\textwidth, height=2.5cm, keepaspectratio]{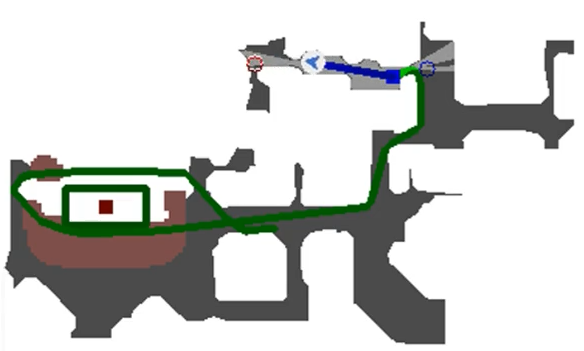} &
        \includegraphics[width=0.19\textwidth, height=2.5cm, keepaspectratio]{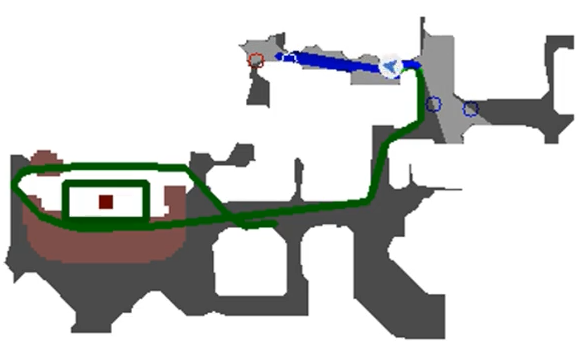} &
        \includegraphics[width=0.19\textwidth, height=2.5cm, keepaspectratio]{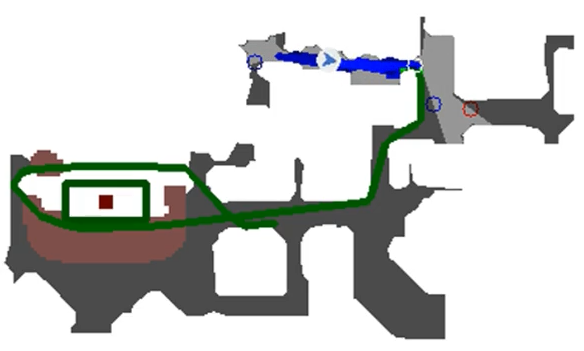} &
        \includegraphics[width=0.19\textwidth, height=2.5cm, keepaspectratio]{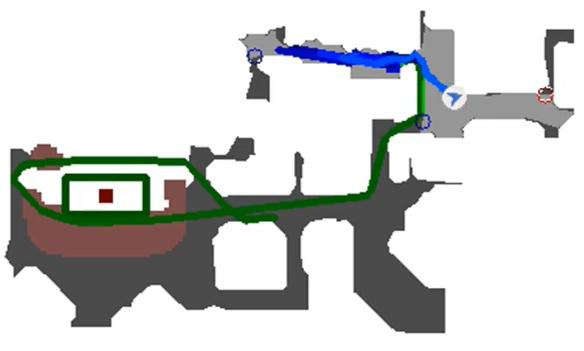} &
        \includegraphics[width=0.19\textwidth, height=2.5cm, keepaspectratio]{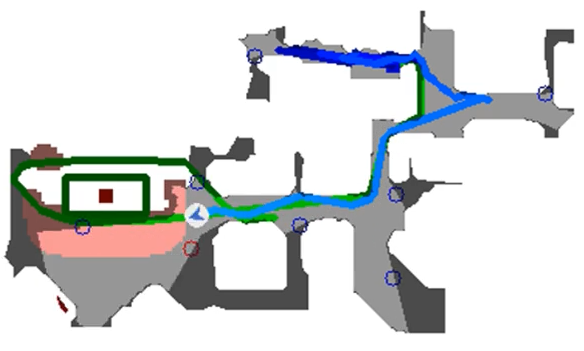} \\[-15pt] % Reduced space after row 2

        % --- ROW 3: Episode 2 (No Action History) ---
        \rotatebox{90}{\parbox{2.5cm}{\centering\small Episode 2 \\ (No History)}} &
        \includegraphics[width=0.19\textwidth, height=2.5cm, keepaspectratio]{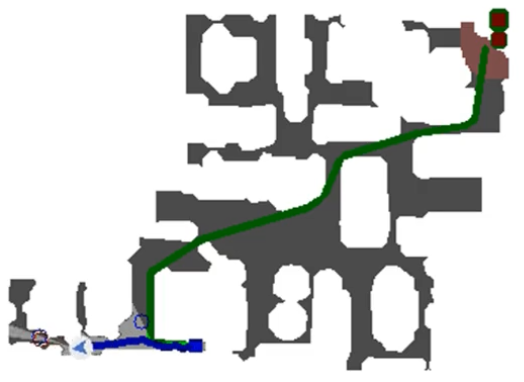} &
        \includegraphics[width=0.19\textwidth, height=2.5cm, keepaspectratio]{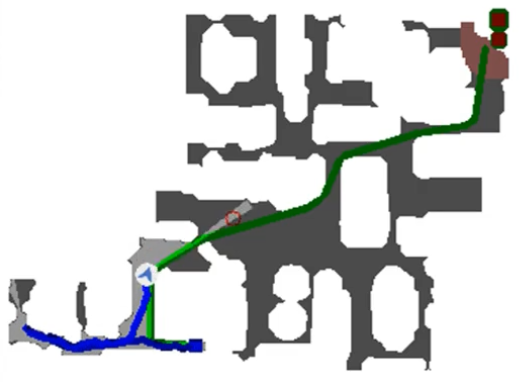} &
        \includegraphics[width=0.19\textwidth, height=2.5cm, keepaspectratio]{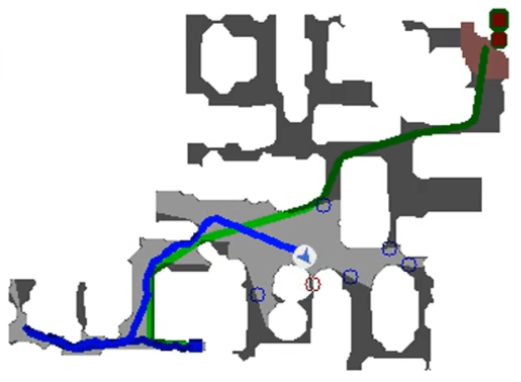} &
        \includegraphics[width=0.19\textwidth, height=2.5cm, keepaspectratio]{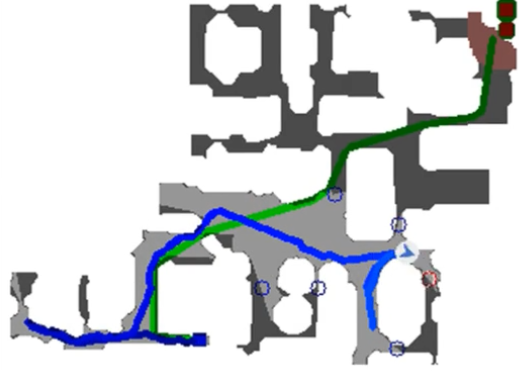} &
        \includegraphics[width=0.19\textwidth, height=2.5cm, keepaspectratio]{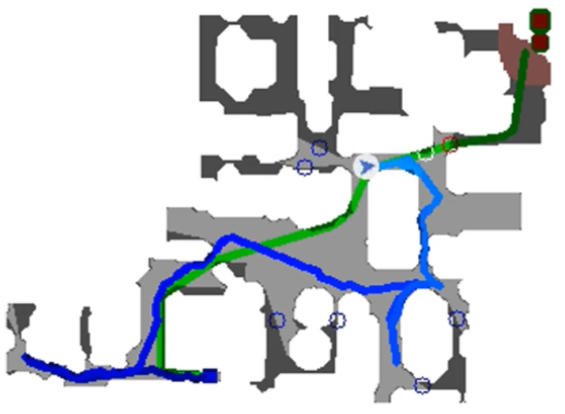} \\[-15pt] % Reduced space after row 3

        % --- ROW 4: Episode 2 (With Action History) ---
        \rotatebox{90}{\parbox{2.5cm}{\centering\small Episode 2 \\ (With History)}} &
        \includegraphics[width=0.19\textwidth, height=2.5cm, keepaspectratio]{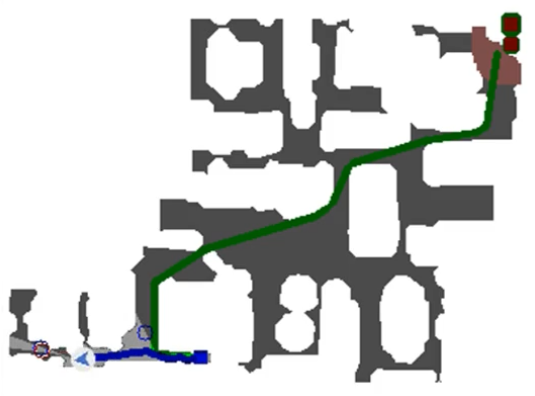} &
        \includegraphics[width=0.19\textwidth, height=2.5cm, keepaspectratio]{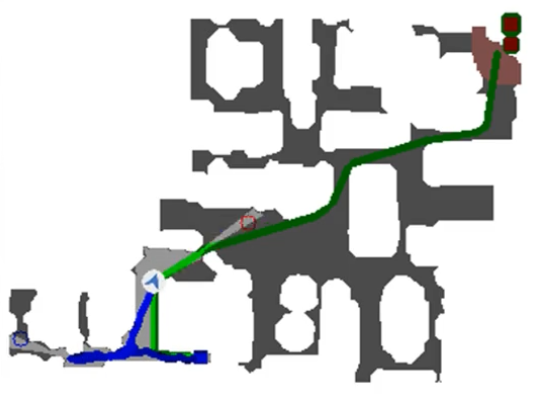} &
        \includegraphics[width=0.19\textwidth, height=2.5cm, keepaspectratio]{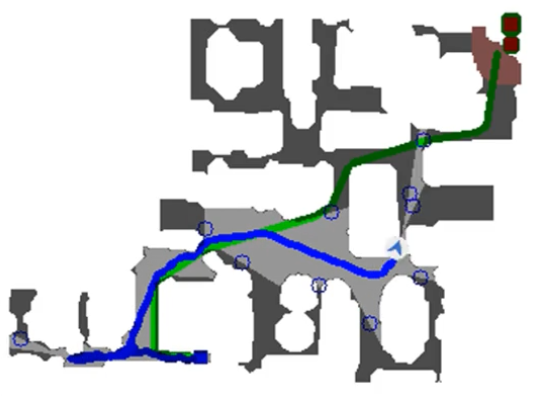} &
        \includegraphics[width=0.19\textwidth, height=2.5cm, keepaspectratio]{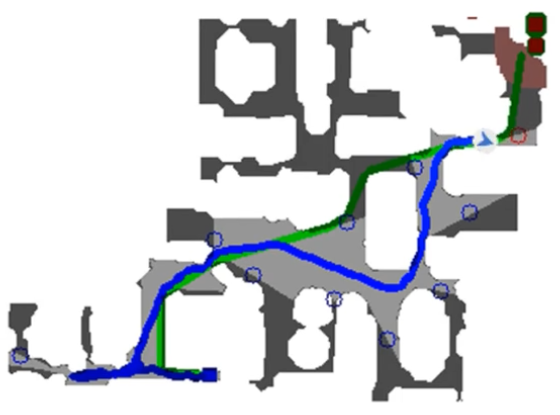} &
        \includegraphics[width=0.19\textwidth, height=2.5cm, keepaspectratio]{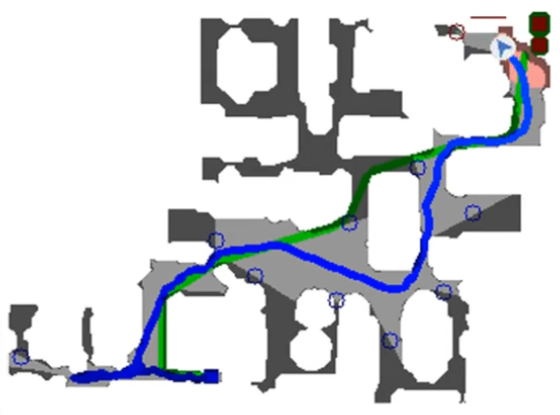} \\
    \end{tabular}
    \caption{Qualitative comparison across two episodes, demonstrating the critical role of the action history module in resolving common navigation failures. 
    \textbf{Episode 1 (Rows 1-2):} The top row shows an agent without action history getting trapped in a corridor. It falls into a persistent decision loop, oscillating between two frontier points and failing to make progress. The second row illustrates how the full framework, using its action history, recognizes the repetitive pattern, breaks the cycle, and successfully finds the target. 
    \textbf{Episode 2 (Rows 3-4):} The third row depicts another failure where the agent without temporal memory gets stuck in indecisive movements, ultimately running out of time. In the final row, the action history module prevents this stagnation, enabling the agent to explore efficiently and reach the goal. Each column shows the agent's view at a specific simulation step.}
    \label{fig:qualitative_history_comparison}
\end{figure*}

The agent without CoT acts impulsively and with little foresight. Its reasoning is basic, connecting a visual stimulus directly to an action without considering the larger situation (e.g., "the TV is on the right... turn right"). This results in aimless wandering and an inability to complete the task on time. In contrast, the agent with our full CoT system follows a more logical and cohesive plan. It methodically analyzes the problem by first identifying the room type ("Bathroom"), then using its general knowledge to determine the probability of finding the target there ("Can a TV be Found Here?: No"), and finally making a sensible decision to search a more likely location ("Recommended Action: Go forward"). This step-by-step approach avoids pointless detours and creates direct and efficient paths. This structured thinking prevents logical inconsistencies in planning. Furthermore, the inclusion of the action history module provides crucial temporal context, allowing the agent to recognize and break out of unproductive loops, as visually demonstrated in Figure~\ref{fig:qualitative_history_comparison}. This ability to avoid stagnation is critical for successful long-horizon navigation.

Our analysis also revealed limitations in the standard evaluation protocol, as we observed episodes marked as failures even when the agent found a valid object of the target category, simply because it was not the specific instance required by the ground truth. A detailed manual analysis of failed episodes provided further insights, categorizing them into five primary modes: annotation incompleteness (25\%), premature episode termination due to loops (24\%), in-view target oversight (19\%), cross-level navigation deficits (18\%), and semantic misclassification (14\%). These findings highlight key areas for improvement in both dataset fidelity and the agent's long-horizon exploration and perception capabilities.

\section{Conclusion and Future Work}
\label{sec:conclusion}
We have presented a new framework for zero-shot object navigation that successfully combines the semantic reasoning of a VLM with methodical frontier-based exploration. Our main contributions—dynamic prompt engineering with CoT reasoning and action history—result in more logical and reliable navigation, achieving state-of-the-art efficiency on widely used benchmarks. Despite these encouraging results, our approach has some limitations that point to areas for future research. The high computational demand of large VLMs is a barrier to real-time use, which calls for advancements in model compression and more efficient architectures. Additionally, our prompt engineering is currently a manual process; automating prompt optimization could improve performance. Future work could also improve the VLM's spatial reasoning by adding more detailed semantic information to the top-down map or by training VLMs specifically on map interpretation.

\FloatBarrier
% --- References ---
\bibliographystyle{plainnat}
\bibliography{references}

%%%%%%%%%%%%%%%%%%%%%%%%%%%%%%%%%%%%%%%%%%%%%%%%%%%%%%%%%%%%
\newpage
\appendix

\section{Appendix}
\label{app:vlm_examples}
This appendix provides the pseudocode for the exploration phase of our approach and the full, optimized VLM prompt that resulted from our iterative refinement process.

\begin{algorithm}[h]
\caption{Object Navigation Exploration Phase}
\label{alg:object_nav_explore}
\begin{algorithmic}[1]
\STATE Initialize map $\mathcal{M}_0$, Value map $\mathcal{V}_0(m) \gets 0.5 \; \forall m$, Confidence $C_0(m) \gets 0 \; \forall m$, pose $p_0$, action history $\mathcal{H}_a \gets \emptyset$
\STATE \textbf{function} Explore($c_{\text{target}}, p_{\text{curr}}, \mathcal{M}_{\text{curr}}, \mathcal{V}_{\text{curr}}, C_{\text{curr}}, \mathcal{H}_{\text{curr}}$)
    \STATE $p_t \gets p_{\text{curr}}; \mathcal{M}_{t-1} \gets \mathcal{M}_{\text{curr}}; \mathcal{V}_{t-1} \gets \mathcal{V}_{\text{curr}}; C_{t-1} \gets C_{\text{curr}}; \mathcal{H}_a \gets \mathcal{H}_{\text{curr}}$
    \FOR{$t = 0$ \textbf{to} MaxExplorationSteps}
        \STATE Get observation $O_t = (I_t, D_t)$ at pose $p_t$
        \STATE $\mathcal{M}_t \gets \text{UpdateMap}(\mathcal{M}_{t-1}, D_t, p_t)$
        \STATE $\mathcal{F}_g \gets \text{FindGeometricFrontiers}(\mathcal{M}_t, p_t)$
        
        \STATE $\pi_{\text{gen},t} \gets \text{GeneratePrompt}(\pi_{\text{template}}, \mathcal{H}_a, c_{\text{target}})$
        \STATE $(S_{\text{act},t}, \beta_{\text{obj},t}, R_{\text{vlm},t}) \gets \text{VLMQuery}(I_t, \mathcal{M}_{t},  \mathcal{H}_a, \pi_{\text{gen},t})$

        \STATE $\mathcal{V}_t \gets \mathcal{V}_{t-1}; C_t \gets C_{t-1}$
        \FOR{each map cell $m$ in current FOV of $O_t$}
            \STATE $v^{\text{vlm}}_m \gets \text{ProjectVLMScore}(S_{\text{act},t},  m, p_t, I_t)$
            \STATE $c^{\text{view}}_m \gets \text{ViewingConfidence}(m, p_t, \text{FOV}_{\text{params}})$ \COMMENT{Eq. \ref{eq:confidence}}
            \STATE $\mathcal{V}_t(m) \gets \frac{c^{\text{view}}_m v^{\text{vlm}}_m + C_{t-1}(m) \mathcal{V}_{t-1}(m)}{c^{\text{view}}_m + C_{t-1}(m) + \epsilon}$ \COMMENT{Eq. \ref{eq:value_update}}
            \STATE $C_t(m) \gets \frac{(c^{\text{view}}_m)^2 + (C_{t-1}(m))^2}{c^{\text{view}}_m + C_{t-1}(m) + \epsilon}$ \COMMENT{Eq. \ref{eq:value_update}}
        \ENDFOR
        
        \STATE $\mathcal{F}_p \gets \emptyset$
        \FOR{each frontier $f \in \mathcal{F}_g$}
            \STATE $v_f \gets \text{QueryValueMap}(\mathcal{V}_t, C_t, f)$
            \STATE Add $(f, v_f)$ to $\mathcal{F}_p$
        \ENDFOR
        \STATE Sort $\mathcal{F}_p$ by $v_f$ descending
        \STATE $w^* \gets \text{SelectExplorationWaypoint}(\mathcal{F}_p)$
        
        \STATE $(p_{t+1}, a_t) \gets \text{MoveTo}(w^*)$
        \STATE $\mathcal{H}_a \gets \text{UpdateActionHistory}(\mathcal{H}_a, a_t)$
        
        \STATE $o_{\text{sensor}} \gets \text{ObjectDetect}(I_t, c_{\text{target}})$
        \IF{$\beta_{\text{obj},t} \textbf{ and } o_{\text{sensor}}$}
            \STATE NavigateToTarget(...) \COMMENT{Transition to goal phase}
            \STATE \textbf{break}
        \ENDIF
    \ENDFOR
\STATE \textbf{end function}
\end{algorithmic}
\end{algorithm}

% \begin{lstlisting}[caption={VLM Response Example}, label={lst:vlm-response}][!t]
% 1. Part of the House: Living room
%     - Reasoning: The image shows a living room area with a large mirror on the wall, which is a common feature in living rooms. The presence of a mirror suggests that the room is designed for socializing and relaxation.

% 2. Can a TV Be Found Here?: Yes
%     - Reasoning: TVs are commonly found in living rooms, as they are used for entertainment purposes. The mirror's presence further supports this being a leisure space.

% 3. Have You Found the TV?: No
%     - Reasoning: The image does not show a television, so it has not been found yet.

% 4. Recommended Action: Go forward
%     - Reasoning: The robot should move forward as there is a higher likelihood of finding a television in this living room area.

% 5. Probability Scores for Each Action:
%     - Go forward: 0.8
%     - Go backward: 0.2
%     - Turn right: 0.1
%     - Turn left: 0.1
%     - Explanation: The probability of finding a television by moving forward is highest (0.8).
% \end{lstlisting}

\begin{lstlisting}[language={}, caption={Example Prompt: No CoT (Action scores only, no reasoning).}, label={lst:prompt_no_cot}]
Robot's Goal: Find the [TARGET_OBJECT].

Provide probability scores for the next navigation action:
- Go forward: [Score between 0.0-1.0]
- Go backward: [Score between 0.0-1.0]
- Turn right: [Score between 0.0-1.0]
- Turn left: [Score between 0.0-1.0]
\end{lstlisting}

\begin{lstlisting}[language={}, caption={Example Prompt: Basic CoT (e.g., "Best action for target?").}, label={lst:prompt_basic_cot}]
You are an autonomous robot navigating an indoor environment.
Your current task is to find a [TARGET_OBJECT].
Image 1 is your current camera view. Image 2 is a top-down map.

Considering your goal and current observations:
1. What is the single best immediate action to take (choose from: go forward, go backward, turn right, turn left) to progress towards finding the [TARGET_OBJECT]?
2. Provide probability scores for all possible actions:
   - Go forward: [Score]
   - Go backward: [Score]
   - Turn right: [Score]
   - Turn left: [Score]

Response Format:
1. Best Action: [Your choice]
2. Probability Scores:
   - Go forward: [Score]
   - Go backward: [Score]
   - Turn right: [Score]
   - Turn left: [Score]
\end{lstlisting}

\begin{lstlisting}[language={}, caption={Example Prompt: Intermediate CoT (e.g., Scene ID + Object Likelihood query).}, label={lst:prompt_intermediate_cot}]
You are a robot assistant searching for a [TARGET_OBJECT] in an indoor environment.
Image 1 shows your current first-person view. Image 2 shows a top-down obstacle map with your current heading indicated by an arrow.

Please analyze the situation and respond by following these steps:
1. Briefly describe the type of area or room you believe you are currently in or about to enter (e.g., kitchen, hallway, living room).
2. Based on the area type and your general knowledge, assess the likelihood (e.g., High, Medium, Low) of finding a [TARGET_OBJECT] in or near this area. Provide a brief justification.
3. Considering your assessment, what is the most logical next navigation action (choose from: go forward, go backward, turn right, turn left)?
4. Provide probability scores (0.0 to 1.0) for each action:
   - Go forward:
   - Go backward:
   - Turn right:
   - Turn left:

Structure your response as follows:
1. Area Type: [Your description]
2. Target Likelihood in Area: [Likelihood (High/Medium/Low)] - Reasoning: [Your justification]
3. Recommended Action: [Your chosen action]
4. Probability Scores for Each Action:
   - Go forward: [Score]
   - Go backward: [Score]
   - Turn right: [Score]
   - Turn left: [Score]
\end{lstlisting}

\begin{lstlisting}[language={}, caption={The optimized prompt after iterative refinement, developed through experimentation with the VLM’s outputs.}, label={lst:prompt_full_cot}]
You are a robot navigating an indoor environment in search of a [TARGET_OBJECT].
Once you find it, move near the [TARGET_OBJECT] and stop.
The first image is your current observation and the second image is a top downview obstacle map of the environment.
The grey areas are obstacles and The robots direction is visible with an arrow.
You must think step by step and ensure that all parts of your response are consistent.

Here are the tasks:
1. Identify what part of the house we are about to enter (choose from: [bedroom, living room, kitchen, corridor, bathroom]).
2. Assess whether a [TARGET_OBJECT] can realistically be found in this area, based on common sense and the current observation.
3. Is there a [TARGET_OBJECT] in the current scene?
4. Determine the most logical next action for the robot (choose from: [go forward, go backward, turn right, turn left]).
- The chosen action must prioritize exploring areas likely to contain a [TARGET_OBJECT].
- Avoid suggesting actions that contradict previous observations (e.g., don't explore a bathroom if couches aren't found there).
- If you are in a corridor, continue your path and Try to exit the corridor and describe where it leads.
- Make Sure the robot isn't stuck in an action loop
5. Provide a probability score for each possible action in the following format:
- Go forward: [Score]
- Go backward: [Score]
- Turn right: [Score]
- Turn left: [Score]

Each probability score should be a number between 0 and 1, with two decimal places of precision.
- A score of 1 means full confidence that the action will lead to finding the [TARGET_OBJECT].
- A score of 0 means no confidence.

When providing your response, use this structure:
1. **Part of the House**: [Your answer]
- Reasoning: [Explain why you think this is the correct part of the house based on the observation and map.]
2. **Can a [TARGET_OBJECT] Be Found Here?**: [Yes/No]
- Reasoning: [Explain why or why not.]
3. **Have You Found the [TARGET_OBJECT]?**: [Yes/No]
4. **Recommended Action**: [Your action]
- Reasoning: [Explain why this action is the most logical based on steps 1 and 2.]
5. **Probability Scores for Each Action**:
- Go forward: [Score]
- Go backward: [Score]
- Turn right: [Score]
- Turn left: [Score]

Important: Ensure that the recommended action aligns with the reasoning from steps 1 and 2. If a [TARGET_OBJECT] cannot be found in the current area, prioritize moving to areas more likely to contain a [TARGET_OBJECT].
\end{lstlisting}

%%%%%%%%%%%%%%%%%%%%%%%%%%%%%%%%%%%%%%%%%%%%%%%%%%%%%%%%%%%%%%%%
% 			 MANDATORY NEURIPS CHECKLIST
%%%%%%%%%%%%%%%%%%%%%%%%%%%%%%%%%%%%%%%%%%%%%%%%%%%%%%%%%%%%%%%%
\newpage
\section*{NeurIPS Paper Checklist}

\begin{enumerate}

\item {\bf Claims}
    \item[] Question: Do the main claims made in the abstract and introduction accurately reflect the paper's contributions and scope?
    \item[] Answer: \answerYes{}
    \item[] Justification: The abstract and introduction claim that our framework improves navigation efficiency (SPL) by deeply integrating a VLM using CoT and action history. This is directly supported by the main results in Table \ref{tab:main_results} and the ablation studies in Table \ref{tab:ablation}.

\item {\bf Limitations}
    \item[] Question: Does the paper discuss the limitations of the work performed by the authors?
    \item[] Answer: \answerYes{}
    \item[] Justification: The conclusion section explicitly discusses limitations, including the computational cost of VLMs and the reliance on manual prompt engineering.

\item {\bf Theory assumptions and proofs}
    \item[] Question: For each theoretical result, does the paper provide the full set of assumptions and a complete (and correct) proof?
    \item[] Answer: \answerNA{}
    \item[] Justification: This paper is empirical and does not present theoretical results or proofs.

\item {\bf Experimental result reproducibility}
    \item[] Question: Does the paper fully disclose all the information needed to reproduce the main experimental results of the paper to the extent that it affects the main claims and/or conclusions of the paper (regardless of whether the code and data are provided or not)?
    \item[] Answer: \answerYes{}
    \item[] Justification: The paper specifies the VLM used (LLaVA-1.6 7B), the datasets (HM3D, MP3D, Gibson), the simulation environment (Habitat), and the core components of the method, including the full prompt in the appendix. This provides a clear path for reproduction.

\item {\bf Open access to data and code}
    \item[] Question: Does the paper provide open access to the data and code, with sufficient instructions to faithfully reproduce the main experimental results, as described in supplemental material?
    \item[] Answer: \answerNo{}
    \item[] Justification: The paper does not include a link to the source code at this time. However, it relies entirely on publicly available datasets and models, and the methodology is described in sufficient detail to allow for independent reimplementation.

\item {\bf Experimental setting/details}
    \item[] Question: Does the paper specify all the training and test details (e.g., data splits, hyperparameters, how they were chosen, type of optimizer, etc.) necessary to understand the results?
    \item[] Answer: \answerYes{}
    \item[] Justification: The paper specifies that it uses the standard validation splits for the HM3D, MP3D, and Gibson datasets as defined by the Habitat Challenge. As a zero-shot method, no training is performed. Key algorithmic details like the decay parameter $\lambda$ and the action history length are provided.

\item {\bf Experiment statistical significance}
    \item[] Question: Does the paper report error bars suitably and correctly defined or other appropriate information about the statistical significance of the experiments?
    \item[] Answer: \answerNo{}
    \item[] Justification: Error bars are not reported. The evaluation is performed over large, standard benchmark datasets containing hundreds to thousands of episodes, and the aggregate metrics (SR, SPL) are standard for this domain, providing a robust measure of performance.

\item {\bf Experiments compute resources}
    \item[] Question: For each experiment, does the paper provide sufficient information on the computer resources (type of compute workers, memory, time of execution) needed to reproduce the experiments?
    \item[] Answer: \answerYes{}
    \item[] Justification: The full paper specifies the VLM model (LLaVA-1.6 7B), and the hardware used (NVIDIA RTX A6000), which provides sufficient information to understand the computational requirements.

\item {\bf Code of ethics}
    \item[] Question: Does the research conducted in the paper conform, in every respect, with the NeurIPS Code of Ethics \url{https://neurips.cc/public/EthicsGuidelines}?
    \item[] Answer: \answerYes{}
    \item[] Justification: The research uses publicly available datasets and models for the task of robotic navigation and does not involve human subjects or sensitive data.

\item {\bf Broader impacts}
    \item[] Question: Does the paper discuss both potential positive societal impacts and negative societal impacts of the work performed?
    \item[] Answer: \answerYes{}
    \item[] Justification: A broader impact statement is included in the conclusion, discussing potential positive applications in assistive robotics and acknowledging potential dual-use concerns.

\item {\bf Safeguards}
    \item[] Question: Does the paper describe safeguards that have been put in place for responsible release of data or models that have a high risk for misuse (e.g., pretrained language models, image generators, or scraped datasets)?
    \item[] Answer: \answerNA{}
    \item[] Justification: The paper does not release new models or datasets. It uses an existing, publicly available VLM. Therefore, safeguards for release are not applicable.

\item {\bf Licenses for existing assets}
    \item[] Question: Are the creators or original owners of assets (e.g., code, data, models), used in the paper, properly credited and are the license and terms of use explicitly mentioned and properly respected?
    \item[] Answer: \answerYes{}
    \item[] Justification: The paper properly cites the creators of the datasets (HM3D, MP3D, Gibson), the simulation environment (Habitat), and the VLM (LLaVA) used. These are standard academic assets with permissive licenses.

\item {\bf New assets}
    \item[] Question: Are new assets introduced in the paper well documented and is the documentation provided alongside the assets?
    \item[] Answer: \answerNA{}
    \item[] Justification: The paper does not introduce any new assets like datasets or models.

\item {\bf Crowdsourcing and research with human subjects}
    \item[] Question: For crowdsourcing experiments and research with human subjects, does the paper include the full text of instructions given to participants and screenshots, if applicable, as well as details about compensation (if any)?
    \item[] Answer: \answerNA{}
    \item[] Justification: This research does not involve crowdsourcing or human subjects.

\item {\bf Institutional review board (IRB) approvals or equivalent for research with human subjects}
    \item[] Question: Does the paper describe potential risks incurred by study participants, whether such risks were disclosed to the subjects, and whether Institutional Review Board (IRB) approvals (or an equivalent approval/review based on the requirements of your country or institution) were obtained?
    \item[] Answer: \answerNA{}
    \item[] Justification: This research does not involve human subjects.

\item {\bf Declaration of LLM usage}
    \item[] Question: Does the paper describe the usage of LLMs if it is an important, original, or non-standard component of the core methods in this research?
    \item[] Answer: \answerYes{}
    \item[] Justification: The Vision-Language Model (a type of LLM) is the central and core methodological component of this research. Its usage is described in detail in Section \ref{sec:method} and \ref{sec:prompt}.

\end{enumerate}
%%%%%%%%%%%%%%%%%%%%%%%%%%%%%%%%%%%%%%%%%%%%%%%%%%%%%%%%%%%%%%%%
\end{document}